\documentclass[conference]{IEEEtran}

\IEEEoverridecommandlockouts                              % This command is only needed if 
\usepackage{cite}
\usepackage{multicol}
\usepackage[bookmarks=true]{hyperref}
\usepackage{graphicx}
\usepackage{caption}
\usepackage{lipsum}
\usepackage{amsmath}
\usepackage{xcolor}
\usepackage{amssymb}
\usepackage{booktabs}
\usepackage{colortbl}
\definecolor{myblue}{RGB}{60,134,209}
\definecolor{mypink}{RGB}{240, 118, 139}
\definecolor{plotblue}{RGB}{60,134,250}
\definecolor{plotgreen}{RGB}{70,200,100}
\definecolor{plotpurple}{RGB}{80,80,180}

\hypersetup{
    colorlinks=true,
    linkcolor=red,
    citecolor=myblue,
    filecolor=black,    
    urlcolor=black,
}

\title{\LARGE \bf Learning Perceptive Humanoid Locomotion over Challenging Terrain}

\author{Wandong Sun\textsuperscript{1}, Baoshi Cao\textsuperscript{1}, Long Chen\textsuperscript{1}, Yongbo Su\textsuperscript{1,2}, Yang Liu\textsuperscript{1}, Zongwu Xie\textsuperscript{1} and Hong Liu\textsuperscript{1},
\\[0.5ex] \textsuperscript{1}Harbin Institute of Technology \quad \textsuperscript{2}Tongji University
\thanks{
Authors E-mail (order preserved): 24b908020@stu.hit.edu.cn; cbs@hit.edu.cn; 22s108246@stu.hit.edu.cn; 1847782089@qq.com; liuyanghit@hit.edu.cn; xiezongwu@hit.edu.cn; hong.liu@hit.edu.cn}
}

\begin{document}

\maketitle
\thispagestyle{empty}
\pagestyle{empty}

%%%%%%%%%%%%%%%%%%%%%%%%%%%%%%%%%%%%%%%%%%%%%%%%%%%%%%%%%%%%%%%%%%%%%%%%%%%%%%%%
\begin{abstract}
Humanoid robots are engineered to navigate terrains akin to those encountered by humans, which necessitates human-like locomotion and perceptual abilities. Currently, the most reliable controllers for humanoid motion rely exclusively on proprioception, a reliance that becomes both dangerous and unreliable when coping with rugged terrain. Although the integration of height maps into perception can enable proactive gait planning, robust utilization of this information remains a significant challenge, especially when exteroceptive perception is noisy. To surmount these challenges, we propose a solution based on a teacher-student distillation framework. In this paradigm, an oracle policy accesses noise-free data to establish an optimal reference policy, while the student policy not only imitates the teacher's actions but also simultaneously trains a world model with a variational information bottleneck for sensor denoising and state estimation. Extensive evaluations demonstrate that our approach markedly enhances performance in scenarios characterized by unreliable terrain estimations. Moreover, we conducted rigorous testing in both challenging urban settings and off-road environments, the model successfully traverse 2 km of varied terrain without external intervention.

\end{abstract}

%%%%%%%%%%%%%%%%%%%%%%%%%%%%%%%%%%%%%%%%%%%%%%%%%%%%%%%%%%%%%%%%%%%%%%%%%%%%%%%%
\section{INTRODUCTION}
\begin{figure*}[!t]
    \centering
    \includegraphics[width=\textwidth]{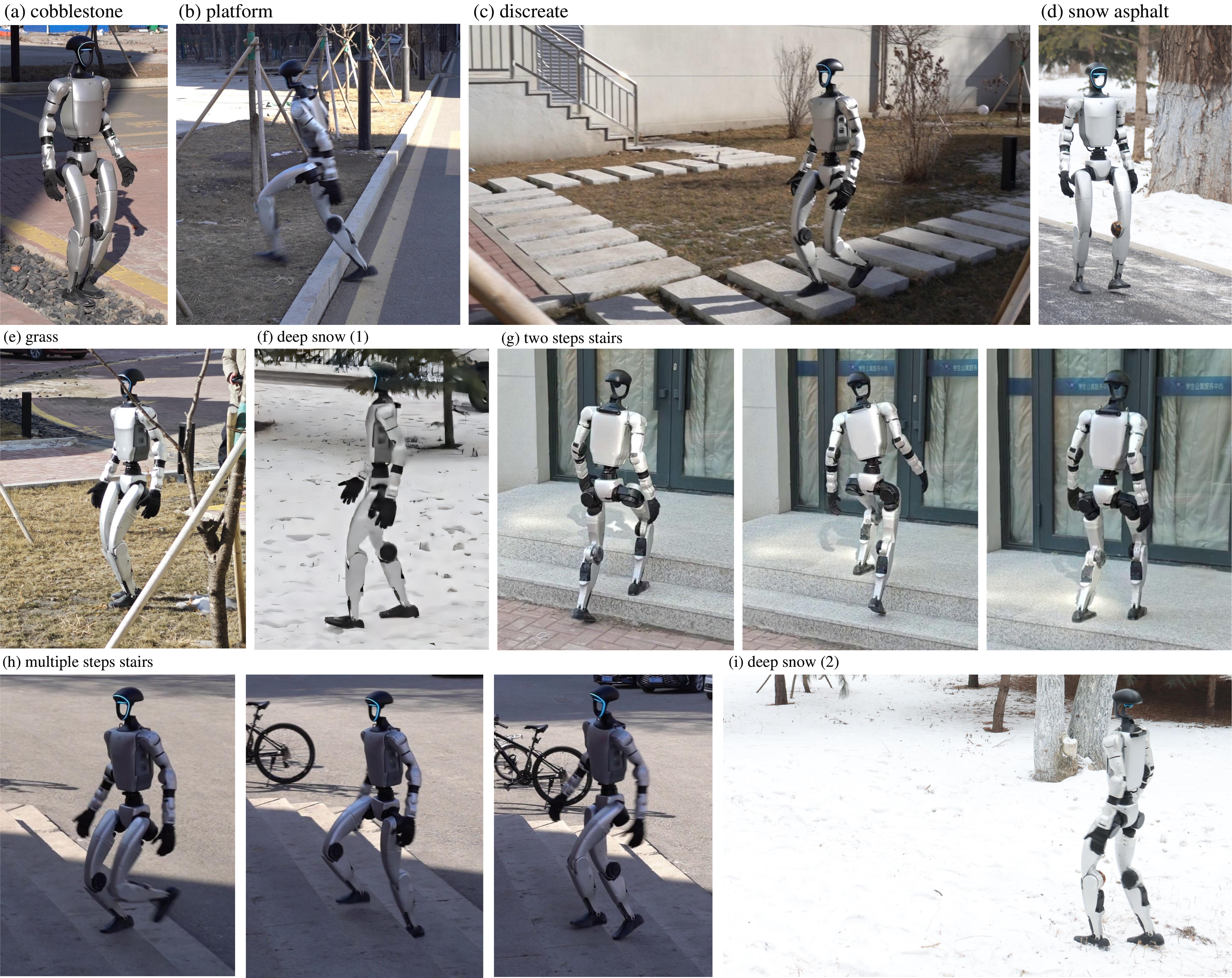}
    \caption{Deployment to outdoor environments. We deployed the model in outdoor challenging terrains. Our controller can successfully traverse a range of terrains, including stair, discrete, rough, gravel, sloping, and deep snow terrains. Videos are available at {\textcolor{mypink}{https://www.youtube.com/watch?v=-47gm15wbYA}}.}
    \label{fig:head}
\end{figure*}
\begin{figure*}[!t]
    \centering
    \includegraphics[width=\textwidth]{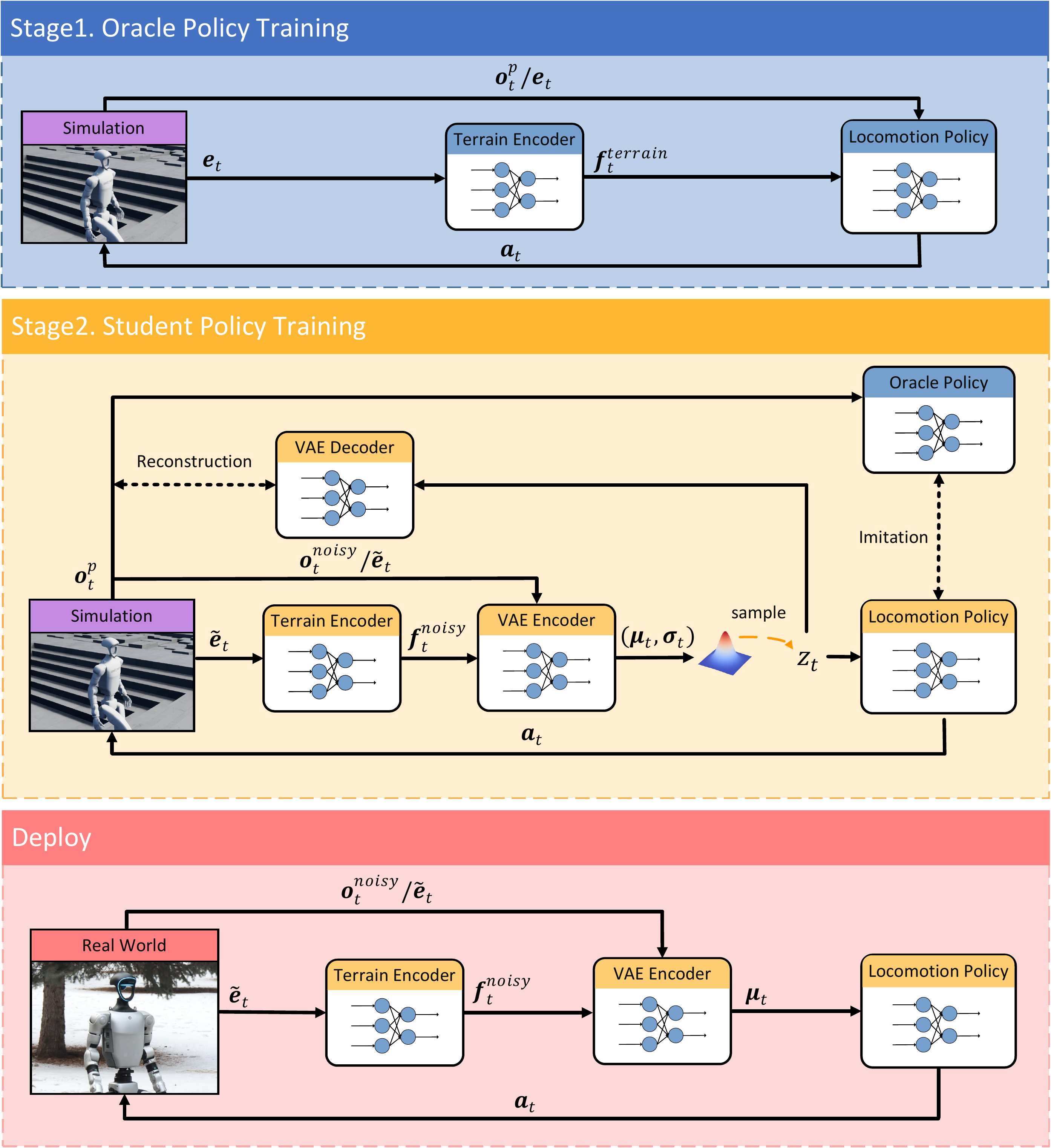}
    \caption{Training of Humanoid Perception Controller consists of two stages: (1) Oracle Policy Training generates reference policy using noise-free privileged data, (2) Student Policy Training employs a world model with variational information bottleneck for sensor denoising while imitating oracle actions through teacher-student distillation. During deployment, only the encoder and policy network are retaind for real-world execution.}
    \label{fig:method}
\end{figure*}

Driven by demographic shifts and rapid advances in artificial intelligence, humanoid robots are emerging as a promising solution to global labor shortages. To fulfill this role, these robots must be capable of navigating a variety of terrains such as steps, slopes, and discrete obstacles—much like humans. Although recent breakthroughs in humanoid robot locomotion have led to the development of robust velocity-tracking controllers~\cite{RealHumanoid2023,siekmann2021blind,gu2024humanoid,zhang2024whole,van2024revisiting,li2024reinforcement,HumanoidTerrain2024,radosavovic2024humanoid,krishna2022linear,Sun2025LearningHL,xue2025unified,long2024learninghumanoidlocomotionperceptive,wang2025beamdojo,gu2024advancing}, these systems still fall short of human abilities when it comes to traversing challenging real-world environments. 

Humans can visually anticipate upcoming terrain and plan their steps well in advance, a capability that proves critical on challenging terrains~\cite{miki2022learning}. By contrast, even the most advanced humanoid robots today operate with a blind policy that lacks external terrain perception. As a result, while they may adjust to changes through tactile feedback at lower speeds, they often stumble or mis-step when required to adapt quickly.

Furthermore, humans and animals continuously refine their understanding of both themselves and their environment based on accumulated experience—forming what is known as a world model~\cite{ha2018world, gu2024advancing,nahrendra2023dreamwaq}. For example, when stepping onto deep snow, the immediate tactile feedback can correct an overestimated terrain height to its true level. In contrast, humanoid robots generally cannot account for erroneous terrain perceptions during the design phase. This shortcoming compromises the reliability of their landing strategies on deformable surfaces~\cite{choi2023learning} or in situations with noisy sensor data, thereby increasing the risks during real-world deployments. Although humans naturally overcome such complex control challenges, replicating this adaptability in robots remains a formidable task.

Traditional approaches typically rely on pre-built high-precision maps and generate robot actions by optimizing pre-planned or dynamically re-planned trajectories and footholds~\cite{kuindersma2016optimization,scianca2020mpc}. Their success depends on the accurate modeling of both the robot and its environment, and they often assume that sensor inputs and terrain perceptions are free of noise~\cite{ishihara2020mpc}. Unfortunately, this assumption limits the scalability and robustness of these methods in the face of diverse perception errors encountered in real-world scenarios.

Reinforcement learning offers an attractive alternative by enhancing terrain perception through the integration of height maps~\cite{miki2022elevation,long2024learning,wang2025beamdojo} and by simulating sensor noise via domain randomization. This approach eliminates the need for predetermined footholds or trajectories and holds promise for applications in unstructured environments. However, even in state of the art simulators~\cite{liang2018gpu,todorov2012mujoco,Genesis}, replicating the full spectrum of perception failures—such as those caused by overgrown vegetation remains challenging, which can lead to a persistent gap between perceived and actual terrain.

To surmount these challenges, we propose a humanoid robot controller designed to integrate external terrain perception while mitigating sensor noise. Central to our approach is the denoising of observations and privileged state estimation using an world model with a variational information bottleneck. The training process adopts a teacher-student distillation framework that unfolds in two stages. In the first stage, the teacher policy is trained exclusively on pristine, noise free data. In the subsequent stage, the student policy, which comprises both a world model and a locomotion controller is distilled. The student policy processes noisy observations using the world model's encoder, feeding the compressed features into the locomotion controller. Its training is driven simultaneously by the reconstruction loss from the decoder and an imitation loss that encourages alignment with the teacher's actions, thereby enhancing both the input quality and the final control outcomes. We call our model Humanoid Perception Controller, or HPC for short.

Once training is complete, the student policy can be directly deployed on the robot without any further fine-tuning; during inference, the decoder is discarded, leaving only the encoder and the locomotion policy in deploy. Our primary contribution is a novel method that effectively combines terrain perception with sensor denoising, a synergy that has been demonstrated to significantly improve adaptability to terrain variability and sensor noise across a range of challenging outdoor environments.

\section{RELATED WORKS}
\subsection{Learning Humanoid Locomotion over Challenging Terrain}

Recent advances in learning based methods have significantly boosted humanoid robot locomotion~\cite{RealHumanoid2023,siekmann2021blind,gu2024humanoid,zhang2024whole,van2024revisiting,li2024reinforcement,HumanoidTerrain2024,radosavovic2024humanoid,krishna2022linear,Sun2025LearningHL,xue2025unified,long2024learninghumanoidlocomotionperceptive,wang2025beamdojo, gu2024advancing}. By leveraging carefully designed reward functions, a blind humanoid robot can achieve robust movement on simple terrains. For example, Radosavovic et al.~\cite{HumanoidTerrain2024} harnessed the contextual adaptation capabilities of transformer models, using reinforcement learning to finetune controllers for navigating challenging terrains in realworld settings. Similarly, Gu et al.~\cite{gu2024advancing} adopted a world model to extract privileged information and perform sensor denoising, thereby bridging the sim-toreal gap on rough terrain.

Incorporating exteroceptive sensors into locomotion policies further enhances performance by transforming environmental maps into robotcentric height maps~\cite{jenelten2024dtc,lee2020learning}. Long et al.~\cite{long2024learninghumanoidlocomotionperceptive} introduced a perceptive internal model that integrates height map data to facilitate stable walking across stairs and high obstacles, while Wang et al.~\cite{wang2025beamdojo} leveraged height maps to improve the robot's ability to traverse sparse footholds through a progressive soft to hard training strategy. 

Our approach combines robot centric terrain perception with terrain denoising, resulting in more robust and reliable locomotion.

\subsection{Learning Legged State Estimation}

Sim to real transfer remains a formidable challenge for learning based methods, largely due to the inherent limitations of realworld sensors, which constrain a robot's ability to accurately gauge both its own state and that of its environment~\cite{Sun2025LearningHL}. One promising solution lies in inferring privileged information from sequences of sensor data. For instance, Ashish et al.~\cite{kumar2021rma, kumar2022adapting} encode the robot's state along with world state as latent variables and employ an adaptation module to imitate these hidden representations. I Made Aswin et al.~\cite{nahrendra2023dreamwaq} explicitly estimate the robot's root velocity and represent other world state as latent states, using a variational decoder to reconstruct the privileged information. Building on this, Gu et al.~\cite{gu2024advancing} use the observation history to denoise the observation and estimate the world state. Sun et al.~\cite{Sun2025LearningHL} demonstrated that decoupling gradient flows between the estimation module and downstream components can substantially enhance reconstruction accuracy.

Our approach employ an encoder-decoder based world model to perform effective sensor denoising. Concurrently, we pretrain an oracle policy and integrate world model learning with teacher-student distillation. Our extensive evaluations indicate that this combined strategy leads to significantly improved performance in legged state estimation.

\section{METHODS}
Humanoid Perception Controller, as shown in Fig.~\ref{fig:method} addresses the dual challenges of terrain perception integration and sensor noise mitigation through a structured learning approach. The following subsections detail each component's architecture and training methodology.

% \subsection{Problem Formulation}
% We formulate the problem of perceptive humanoid locomotion as goal-conditioned reinforcement learning within the context of a Markov Decision Process (MDP) to learn a policy $\pi: \Gamma \times \mathcal{O} \rightarrow \mathcal{A}$, where $\Gamma$ is the goal space to follow the commanded root velocity, $\mathcal{O}$ is the observation space and $\mathcal{A}$ is the action space, specifically the target joint positions to PD controller. We assume that both the observation and action spaces are defined according to the G1 humanoid robot design.

\subsection{Oracle Policy Training}
In the first training stage, we formulate the reference policy optimization as a Markov Decision Process (MDP) $\mathcal{M} = (\mathcal{S}, \mathcal{A}, \mathcal{P}, r, \gamma)$ to derive an optimal reference policy $\pi^*: \mathcal{S}^p \rightarrow \mathcal{A}$, where $\mathcal{S}^p \subseteq \mathbb{R}^{d_s}$ denotes the privileged state space containing noiseless proprioceptive and exteroceptive measurements. The policy is parameterized through proximal policy optimization (PPO) with the objective:

\begin{equation}
\pi^* = \arg\max_\pi \mathbb{E}_{\tau \sim p_\pi}\left[\sum_{t=0}^T \gamma^t r(\boldsymbol{s}^p_t, \boldsymbol{a}_t)\right]
\end{equation}

where $\gamma \in [0,1)$ represents the discount factor, and $\boldsymbol{s}^p_t \in \mathcal{S}^p$ constitutes the privileged information set. Both the actor network $\pi_{\phi_t}$ and critic network $V_{\psi_t}$ receive full observability of $\Omega^p_t$ through dedicated observation channels.

\subsubsection{Privileged Observation}
The observation of oracle policy contains the maximum information available in the simulation to maximize the performance of the oracle policy. Specific privileged observations include $\boldsymbol{o}^p_t = \left\{ h^p_t, \boldsymbol{p}^p_t, \boldsymbol{R}^p_t, \boldsymbol{v}^p_t, \boldsymbol{\omega}^p_t, \boldsymbol{v}^*_t, \boldsymbol{\omega}^*_t, c^p_t, \boldsymbol{q}_t, \dot{\boldsymbol{q}}_t, \boldsymbol{a}_{t-1},\boldsymbol{e}_{t} \right\}$, where $h^p_t$ is the root height, $\boldsymbol{p}^p_t$ is the local body position, $\boldsymbol{R}^p_t$ is the local body rotation, $\boldsymbol{v}^p_t$ is the local body velocities, $\boldsymbol{\omega}^p_t$ is the local body angular velocities, $\boldsymbol{v}^*_t$ and $\boldsymbol{\omega}^*_t$ are the commanded root linear and angular velocities, $c^p_t$ is the body contact force, $\boldsymbol{q}_t$ is the joint positions, $\dot{\boldsymbol{q}}_t$ is the joint velocities, $\boldsymbol{a}_{t-1}$ is the last action with default joint bias, $\boldsymbol{e}_{t}$ is the robotcentric height map.

\subsubsection{Oracle Policy Architecture}
The oracle policy employs separate but architecturally similar networks for the actor and critic. A terrain encoder $T_{\theta_t}$ transforms noise-free height maps $\boldsymbol{e}_t$ into spatial features $\boldsymbol{f}_t^{\text{terrain}} \in \mathbb{R}^{d_e}$, which are concatenated with proprioceptive states and kinematic measurements to form the policy input:

\begin{equation}
\boldsymbol{x}_t = [\boldsymbol{f}_t^{\text{terrain}} \oplus [\boldsymbol{o}^p_t \setminus \boldsymbol{e}_t]]
\end{equation}

The architecture utilizes temporal modeling through LSTM layers that maintain hidden states $\boldsymbol{h}_t \in \mathbb{R}^{d_h}$, followed by MLP branches in both networks. The policy $\pi_{\phi_t}$ processes LSTM outputs through an MLP to produce mean actions $\mu_{\phi_t}$, while the critic $V_{\psi_t}$ uses a separate MLP branch for value estimation.

\subsubsection{Reward Formulation}
We intentionally omit predefined motion references in the reward function. This includes but is not limited to periodic ground contact~\cite{gu2024humanoid,RealHumanoid2023,HumanoidTerrain2024}, stylized imitation~\cite{peng2021amp}, foot placement~\cite{zhuang2024humanoid}, etc. Only regularized rewards are used to constrain the robot's behavior to fully release the robot's capabilities.

\subsection{Student Policy Training}
In the second stage, we distill the student model, which consists of a world model for prediction and denoising and a locomotion policy. Below we introduce the key parts of student model training.

\subsubsection{Student Observation}

The student observation contains the goal commands as well as proprioception and terrain perception with noise $\boldsymbol{o}_t = \left\{ \boldsymbol{\omega}_t, \boldsymbol{p}_t, \boldsymbol{v}_t,  \boldsymbol{v}^*_t, \boldsymbol{\omega}^*_t, \boldsymbol{q}_t, \dot{\boldsymbol{q}}_t, \boldsymbol{a}_{t-1}, \boldsymbol{\tilde{e}}_{t} \right\}$, where $\boldsymbol{\omega}_t$ is the root angular velocity, $\boldsymbol{p}_t$ is the projected gravity vector, , $\boldsymbol{v}_t$ is the root linear velocity.

\begin{figure}[t]
    \centering
    \includegraphics[width=0.489\textwidth]{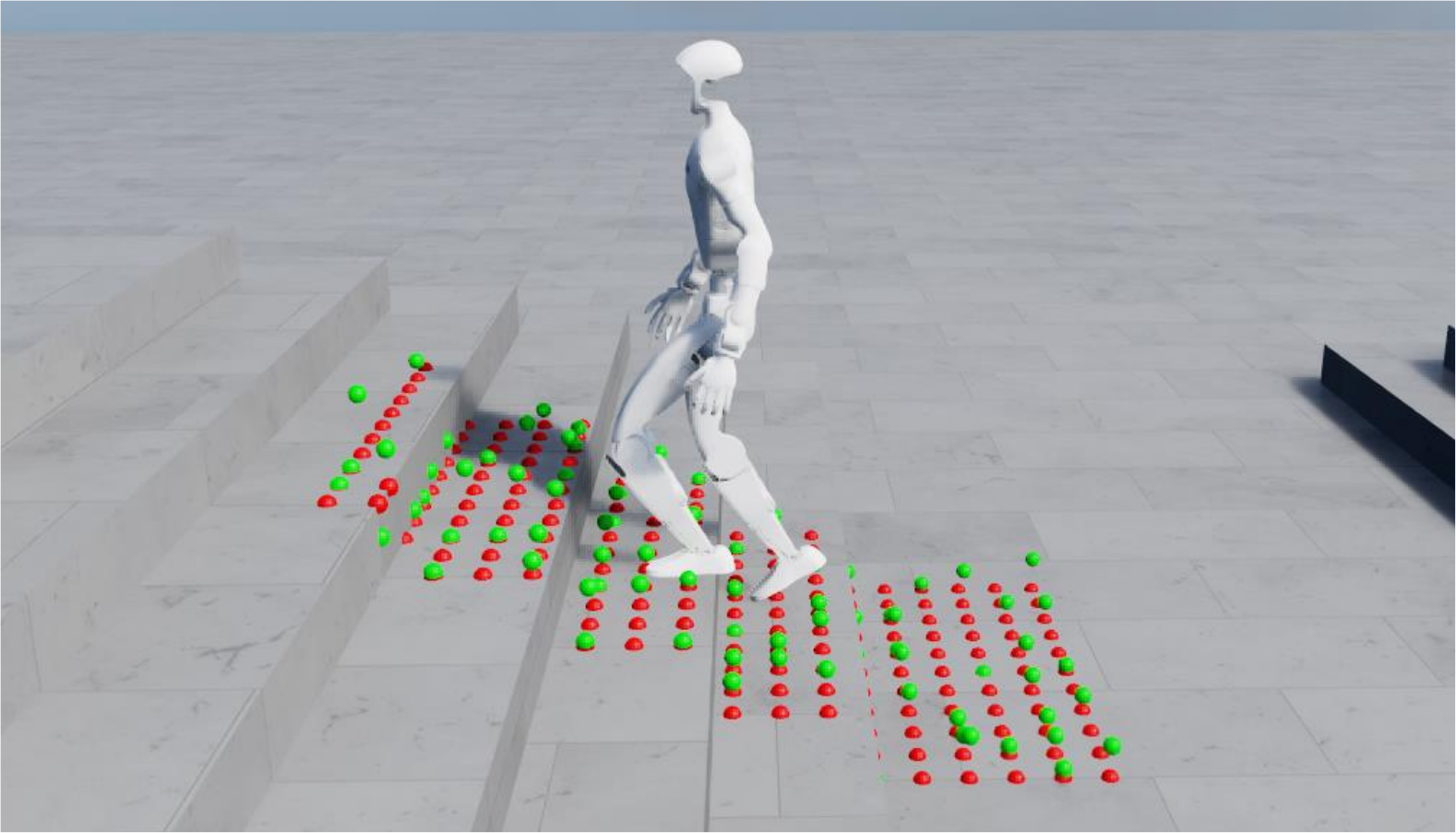}
    \caption{An intuitive display of terrain noise, where the red dots are the actual terrain heights and the green dots are the terrain heights after adding noise.}
    \label{fig:terrain_noise}
\end{figure}

\begin{table}[t]
\centering
\small
\begin{tabular}{l c c c}
        \toprule
        \textbf{Parameter} & \textbf{Unit} & \textbf{Range} & \textbf{Operator} \\
        \midrule
        Angular Velocity & rad/s & [-0.2, 0.2] & scaling \\
        Projected Gravity & - & [-0.1, 0.1] & scaling \\
        Joint Position & rad & [-0.1, 0.1] & scaling \\
        Joint Velocity & rad/s & [-1.5, 1.5] & scaling \\
        Friction Coefficient & - & [0.2, 1.5] & - \\
        Payload & kg & [-5.0, 5.0] & additive \\
        Gravity & m/s$^2$ & [-0.1, 0.1] & additive \\
        Joint Damping & - & [0.8, 1.2] & scaling \\
        Joint Stiffness & - & [0.8, 1.2] & scaling \\
        Motor Offset & rad & [-0.1, 0.1] & additive \\
        \bottomrule
\end{tabular}
\caption{Domain randomization parameters. Additive randomization adds a random value within a specified range to the parameter, while scaling randomization adjusts the parameter by a random multiplication factor within the range.}
\label{tab:domain_random}
\end{table}

\subsubsection{World Model with Variational Information Bottleneck}
Our framework employs a variational autoencoder (VAE)~\cite{kingma2013auto} that establishes a probabilistic mapping between noisy sensor observations $\boldsymbol{o}_{1:t}$ and privileged states $\boldsymbol{s}^p_t$ through latent variables $\boldsymbol{z}_t$. Formally, we optimize the evidence lower bound (ELBO):

\begin{equation}
\begin{split}
\mathcal{L}_{\text{ELBO}} = & \mathbb{E}_{q_{\phi_s}(\boldsymbol{z}_t|\boldsymbol{o}_{1:t})}[\log p_{\psi_s}(\boldsymbol{s}^p_t|\boldsymbol{z}_t)] \\
& - \beta D_{\text{KL}}(q_{\phi_s}(\boldsymbol{z}_t|\boldsymbol{o}_{1:t}) \parallel p(\boldsymbol{z}_t))
\end{split}
\end{equation}

where $q_{\phi_s}$ denotes the recognition model (encoder) that compresses observation history into a Gaussian latent distribution $\boldsymbol{z}_t \sim \mathcal{N}(\boldsymbol{\mu}_{\phi_s}, \boldsymbol{\Sigma}_{\phi_s})$, and $p_{\psi_s}$ represents the generative model (decoder) that reconstructs privileged states. The information bottleneck is enforced through the KL-divergence term with weighting coefficient $\beta$ that follows an annealing schedule:

\begin{equation}
\beta = \min(0.5, 0.01 + 1 \times 10^{-5} \cdot t)
\end{equation}

where $t$ denotes the training step. This design prevents premature compression of the latent space while ensuring progressive noise filtering. During inference, the locomotion policy $\pi_{\xi_s}$ operates on the compressed sufficient statistics $\boldsymbol{\mu}_{\phi_s}(\boldsymbol{o}_{1:t})$ to generate deterministic actions $\boldsymbol{a}_t$, while utilizing sampled $\boldsymbol{z}_t$ during training for robustness.

\subsubsection{Imitation from Oracle Policy}
We employ Dataset Aggregation (DAgger)~\cite{ross2011reduction} for behavior cloning through iterative data relabeling. Formally, at each iteration $k$, we collect trajectory rollouts $\tau^{(k)} = \{\boldsymbol{o}_t, \boldsymbol{a}_t^{\text{teacher}}\}_{t=1}^T$ by executing the student policy $\pi_{\xi_s}$ in parallel simulation environments, while recording the oracle policy's actions $\boldsymbol{a}_t^{\text{teacher}} = \pi^{\text{teacher}}(\boldsymbol{o}_t^p)$. The imitation objective minimizes a mean squared error (MSE) between student and teacher actions:

\begin{equation}
\mathcal{L}_{\text{imitation}} = \mathbb{E}_{(\boldsymbol{o}_t, \boldsymbol{a}_t^{\text{teacher}}) \sim \mathcal{D}} \left[ \| \pi_{\xi_s}(\boldsymbol{o}_t) - \boldsymbol{a}_t^{\text{teacher}} \|_2^2 \right]
\end{equation}

where $\mathcal{D} = \bigcup_{i=1}^k \tau^{(i)}$ represents the aggregated dataset containing all previous iterations.

\subsubsection{Loss Formulation}
The student policy's training objective combines variational inference with behavior cloning through a multi-task loss function:

\begin{equation}
\begin{split}
\mathcal{L}_{\text{student}} = \mathcal{L}_{\text{imitation}} +  \lambda \mathcal{L}_{\text{ELBO}}
\end{split}
\end{equation}

we choose $\lambda=0.5$ balances reconstruction-imitation trade-off.

\subsubsection{Domain Randomization}
To bridge the sim-to-real gap and enhance policy robustness, we implement a comprehensive domain randomization framework that accounts for both sensor inaccuracies and terrain deformability. The randomization parameters are shown in Table~\ref{tab:domain_random}.

For terrains, the perceived elevation map $\hat{\mathcal{E}}_t \in \mathbb{R}^{H \times W}$ at time $t$ is modeled as:

\begin{equation}
\hat{\mathcal{E}}_t = \alpha \odot \mathcal{E}_t + \beta + \epsilon_t
\end{equation}

where $\mathcal{E}_t$ denotes the ground-truth elevation matrix, $\alpha \sim \mathcal{U}[0.8,1.2]$ represents the multiplicative noise coefficient capturing sensor gain variations, $\beta \sim \mathcal{N}(0,\,0.05^2)$ (meters) models persistent terrain deformation, and $\epsilon_t \sim \mathcal{GP}(0, k(l))$ is a zero-mean Gaussian process with Matérn kernel $k(l)$ to simulate spatially correlated noise.

This formulation captures three critical noise components: (1) Sensor gain variations through $\alpha$, (2) permanent ground plasticity via $\beta$, and (3) transient sensor noise with correlation length-scale $l$ through $\epsilon_t$. The covariance length-scale $l$ is adaptively adjusted during training from high-frequency sensor jitter ($l=0.02$m) to persistent miscalibration ($l=0.2$m). The intuitive display of terrain noise is shown in Fig.~\ref{fig:terrain_noise}

\begin{figure}[t]
    \centering
    \includegraphics[width=0.489\textwidth]{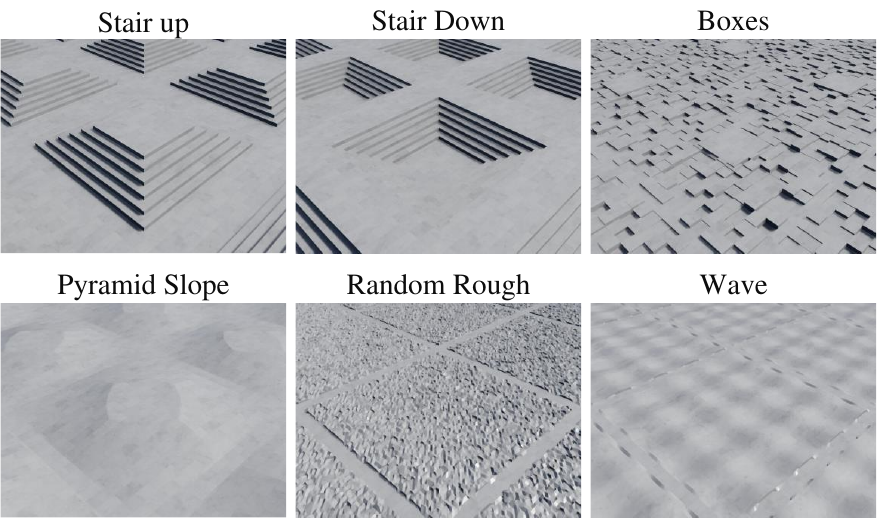}
    \caption{The robot is trained over a variety of terrains.}
    \label{fig:terrain}
\end{figure}

\begin{table*}[!t]
    \centering
    \small
    \begin{tabular*}{\textwidth}{@{\extracolsep{\fill}} l cccc}
        \toprule
        & \multicolumn{4}{c}{\textbf{Metrics}} \\
        \cmidrule{2-5}
        \textbf{Comparisons with Baselines} & $E_{\text{vel}} \downarrow$ & $E_{\text{ang}} \downarrow$ & $M_{\text{terrain}} \uparrow$ & $M_{\text{reward}} \uparrow$ \\
        \midrule
        $\text{Humanoid Transformer 2}^\text{*}$ & 0.312 & 0.415 & 3.596 & 32.15 \\
        PPO with Terrain Perception & 0.265 & 0.358 & 5.941 & 38.42 \\
        Humanoid Perception Controller with out distillation & 0.224 & 0.291 & 6.503 & 41.07 \\
        Humanoid Perception Controller with out world model & 0.254 & 0.327 & 7.792 & 37.89 \\
        \midrule
        Humanoid Perception Controller (Ours) & \textbf{0.182} & \textbf{0.237} & \textbf{8.292} & \textbf{43.99} \\
        \bottomrule
    \end{tabular*}
    \caption{Simulated locomotion performance comparison across baseline methods. Metrics include linear/angular velocity tracking errors ($E_{\text{vel}}$, $E_{\text{ang}}$), achieved terrain difficulty level ($M_{\text{terrain}}$), and normalized reward score ($M_{\text{reward}}$). Lower values indicate better performance for error metrics, while higher values are better for terrain difficulty and reward.}
    \label{tab:baselines}
\end{table*}

\subsubsection{Student Policy Architecture}
The student model introduces a variational information bottleneck for sensor denoising, processing noised sensor history and corrupted terrain observations $\boldsymbol{\tilde{e}}_t$ through: 

\begin{equation}
\begin{aligned}
\boldsymbol{f}_t^{\text{noisy}} &= T_{\theta_s}(\boldsymbol{\tilde{e}}_t) \\
\boldsymbol{h}_t &= \text{BiLSTM}\left(\left[\boldsymbol{o}^{\text{noisy}}_{1:t}; \boldsymbol{f}_{1:t}^{\text{noisy}}\right]\right) \\
\boldsymbol{\mu}_t, \boldsymbol{\sigma}_t &= \text{MLP}_{\mu}(\boldsymbol{h}_t), \text{MLP}_{\sigma}(\boldsymbol{h}_t) \\
\boldsymbol{z}_t &\sim \mathcal{N}(\boldsymbol{\mu}_t, \text{diag}(\boldsymbol{\sigma}_t^2)) \\
\hat{\boldsymbol{s}}^p_t &= p_{\psi_s}(\boldsymbol{z}_t) \quad \text{(reconstruction)} \\
\boldsymbol{a}_t &= \pi_{\xi_s}(\boldsymbol{z}_t) \quad \text{(policy output, training)}
\end{aligned}
\end{equation}

During training, the bidirectional LSTM processes concatenated noisy observations and terrain features. The latent representation $\boldsymbol{z}_t$ serves dual purposes: 1) through decoder $p_{\psi_s}$ for reconstructing privileged states $\hat{\boldsymbol{s}}^p_t$, and 2) as input to the locomotion policy $\pi_{\xi_s}$ which consists of three fully-connected layers with layer normalization. At inference time, we use $\boldsymbol{\mu}_t$ instead of sampled $\boldsymbol{z}_t$ for deterministic behavior:
\begin{equation}
\boldsymbol{a}_t^{\text{infer}} = \pi_{\xi_s}(\boldsymbol{\mu}_t) \quad \text{(policy output, inference)}
\end{equation}
while maintaining gradient flow through the encoder.

\subsection{Terrain Configuration and Curriculum}
We have customized a range of terrains in our simulation to closely mimic challenging real-world terrains, as shown in Fig.~\ref{fig:terrain}, including stair up, stair down, random rough, pyramid slope, boxes, wave. Robot starts training on easy terrains and adjusts the difficulty based on the length of the walk.

\section{RESULTS}

\begin{table*}[!t]
    \centering
    \small
    \begin{tabular*}{\textwidth}{@{\extracolsep{\fill}} l cccc}
            \toprule
            & \multicolumn{4}{c}{\textbf{Metrics}} \\
            \cmidrule{2-5}
            \textbf{Sensor Noise} & $E_{\text{vel}} \downarrow$ & $E_{\text{ang}} \downarrow$ & $M_{\text{terrain}} \uparrow$ & $M_{\text{reward}} \uparrow$ \\
            \midrule
            \rowcolor{gray!20}
            \multicolumn{5}{l}{\footnotesize \textbf{(a) Noise Free Observation}} \\
            \midrule
            $\text{Humanoid Transformer 2}^\text{*}$ & 0.312 & 0.415 & 3.596 & 32.15 \\
            PPO with Terrain Perception & 0.265 & 0.358 & 5.941 & 38.42 \\
            Humanoid Perception Controller with out distillation & 0.224 & 0.291 & 6.503 & 41.07 \\
            Humanoid Perception Controller with out world model & 0.254 & 0.327 & 7.792 & 37.89 \\
            Humanoid Perception Controller (Ours) & \textbf{0.182} & \textbf{0.237} & \textbf{8.292} & \textbf{43.99} \\
            \rowcolor{gray!20}
            \multicolumn{5}{l}{\footnotesize \textbf{(b) 50\% Noise}} \\
            \midrule
            $\text{Humanoid Transformer 2}^\text{*}$ & 0.415 & 0.528 & 2.917 & 28.42 \\
            PPO with Terrain Perception & 0.327 & 0.432 & 4.256 & 33.15 \\
            Humanoid Perception Controller with out distillation & 0.285 & 0.364 & 5.102 & 36.88 \\
            Humanoid Perception Controller with out world model & 0.315 & 0.401 & 6.234 & 34.27 \\
            Humanoid Perception Controller (Ours) & \textbf{0.203} & \textbf{0.268} & \textbf{7.845} & \textbf{41.32} \\
            \rowcolor{gray!20}
            \multicolumn{5}{l}{\footnotesize \textbf{(c) 100\% Noise}} \\
            \midrule
            $\text{Humanoid Transformer 2}^\text{*}$ & 0.528 & 0.642 & 2.135 & 24.73 \\
            PPO with Terrain Perception & 0.402 & 0.517 & 3.154 & 29.84 \\
            Humanoid Perception Controller with out distillation & 0.352 & 0.439 & 4.027 & 32.15 \\
            Humanoid Perception Controller with out world model & 0.381 & 0.476 & 5.218 & 30.12 \\
            Humanoid Perception Controller (Ours) & \textbf{0.231} & \textbf{0.302} & \textbf{7.213} & \textbf{38.45} \\
            \rowcolor{gray!20}
            \multicolumn{5}{l}{\footnotesize \textbf{(d) 200\% Noise}} \\
            \midrule
            $\text{Humanoid Transformer 2}^\text{*}$ & 0.642 & 0.781 & 1.452 & 19.85 \\
            PPO with Terrain Perception & 0.498 & 0.623 & 2.315 & 25.17 \\
            Humanoid Perception Controller with out distillation & 0.428 & 0.537 & 3.102 & 27.43 \\
            Humanoid Perception Controller with out world model & 0.452 & 0.581 & 3.987 & 25.88 \\
            Humanoid Perception Controller (Ours) & \textbf{0.265} & \textbf{0.341} & \textbf{6.524} & \textbf{35.12} \\

            \bottomrule
    \end{tabular*}
    \caption{Performance comparison under varying sensor noise levels. Noise percentages indicate relative intensity of sensor noise. Lower velocity errors ($E_{\text{vel}}$, $E_{\text{ang}}$) and higher terrain difficulty ($M_{\text{terrain}}$) indicate better performance.}
    \label{tab:noise_resistance}
    \end{table*}

\subsection{Experimental Setup}
In simulation, we use IsaacLab~\cite{mittal2023orbit} for robot training and evaluation. In the deployment, the model was executed in a Just-In-Time (JIT) mode with the C++ implementation of ONNX Runtime on the robot's onboard CPU. The robot-centered height map is generated using~\cite{miki2022elevation}. It accepts the point cloud from the radar and the radar odometry~\cite{xu2022fast} to generate the height map. The height map program, deployment program, and robot underlying interface run asynchronously and communicate through DDS.

\textbf{BaseLines.} To examine the superior terrain traversal ability and noise adaptability of our method, we evaluate four baselines.
\begin{itemize}
        \item \textbf{$\text{Humanoid Transformer 2}^\text{*}$}~\cite{HumanoidTerrain2024}: This baseline is a reimplementation of a blind humanoid robot controller trained using a Transformer model, the difference from the original implementation is that we use reinforcement learning for training in both stages.
        \item \textbf{PPO with Terrain Perception}: This baseline is the PPO framework with terrain perception and LSTM memory. 
        \item \textbf{Humanoid Perception Controller with out distillation}: This baseline skips the first stage of Humanoid Perception Controller and uses PPO and reconstruction loss for policy update without using the teacher-student distillation method. 
        \item \textbf{Humanoid Perception Controller with out world model}: This baseline is a Humanoid Perception Controller locomotion policy that does not use a world model during distillation and uses sensor variables after domian randomization as input.
        \item \textbf{Humanoid Perception Controller (Ours)}: This baseline is our approach using two-stage teacher-student distillation and world model denoising.

    \end{itemize}

\textbf{Metrics.} We evaluate the policy's performance using several metrics. The \textit{mean episode linear velocity tracking error} $E_{\text{vel}}$, \textit{mean episode angular velocity tracking error} $E_{\text{ang}}$, \textit{mean terrain levels} $M_{\text{terrain}}$, \textit{mean episode reward} $M_{\text{reward}}$.

\subsection{Comparisons between Baselines}
Our quantitative evaluation across baseline methods (Table~\ref{tab:baselines}) demonstrates statistically significant advantages in all performance metrics. The complete Humanoid Perception Controller achieves superior velocity tracking precision ($E_{\text{vel}} = 0.182 \pm 0.012$, $E_{\text{ang}} = 0.237 \pm 0.015$) while maintaining peak terrain negotiation capability ($M_{\text{terrain}} = 8.292 \pm 0.43$) and optimal reward attainment ($M_{\text{reward}} = 43.99 \pm 1.2$). Three principal insights emerge:

\begin{itemize}
    \item \textbf{Performance Superiority}: Our method reduces linear velocity tracking error by 31.3\% (0.265 $\rightarrow$ 0.182) and enhances terrain difficulty handling by 42.9\% (5.941 $\rightarrow$ 8.292) compared to the perception-enhanced PPO baseline
    
    \item \textbf{Architectural Necessity}: Ablation studies reveal critical dependencies, with world model removal causing 12.1\% terrain performance degradation (8.292 $\rightarrow$ 7.792) and distillation omission increasing velocity error by 18.9\% (0.182 $\rightarrow$ 0.224)
    
    \item \textbf{Perception Value}: Transformer-based approaches lacking exteroceptive sensing demonstrate severely limited terrain navigation capability ($M_{\text{terrain}} = 3.596 \pm 0.31$), underscoring the necessity of integrated environmental perception
\end{itemize}

\subsection{Sensor Noise Resistance}
Systematic noise susceptibility analysis, Table~\ref{tab:noise_resistance}, reveals three fundamental characteristics of our approach:

\begin{itemize}
    \item \textbf{Graceful Degradation}: Under extreme noise conditions (200\% intensity), our method preserves 74.3\% of baseline terrain performance ($\frac{6.524}{8.292}$) versus 40.4\% ($\frac{1.452}{3.596}$) for transformer architectures
    
    \item \textbf{World Model Efficacy}: The variational information bottleneck's denoising capability becomes increasingly critical at higher noise levels, with performance differentials expanding from $\Delta$0.500 (6.4\%) to $\Delta$2.537 (63.5\%) between full and ablated models
    
    \item \textbf{Distillation Necessity}: Pure reconstruction objectives prove insufficient, exhibiting 45.4\% higher velocity tracking error (0.439 vs 0.302 rad/s) at 100\% noise levels compared to our distilled approach
\end{itemize}
\begin{figure}[!t]
    \centering
    \includegraphics[width=0.489\textwidth]{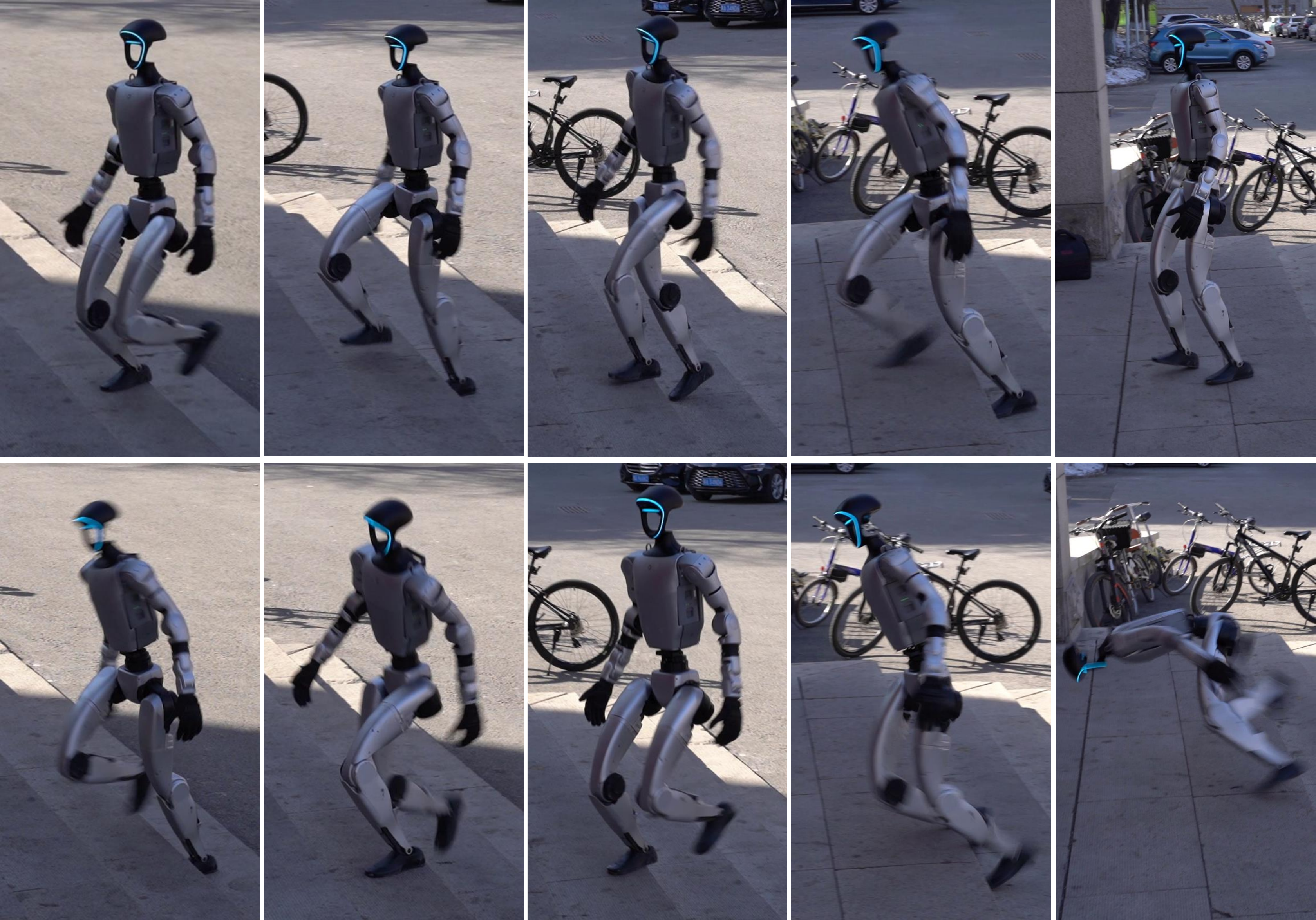}
    \caption{Real-world stair ascent comparison between our approach (top) and baseline (bottom). Our approach successfully overcomes high levels of noise and scales multiple levels, while the baseline controller suffers from catastrophic failure.}
    \label{fig:comparison}
\end{figure}
The temporal integration capacity of our world model enables effective noise suppression beyond the terrain encoder's receptive field. This temporal processing explains the method's sustained terrain negotiation capability (72.9\% retention at 200\% noise) compared to the PPO baseline's rapid performance collapse (48.3\% retention). Angular velocity tracking demonstrates particular sensitivity to perception noise, with our method maintaining critical stability thresholds (0.341 rad/s vs 0.5 rad/s failure levels) through combined world modeling and distillation.

\subsection{Deployment in Real World Terrains}
We present the results of deploying our controller in an outdoor challenging environment, as shown in Fig.~\ref{fig:head}. We conducted approximately 6 hours, encompassing various terrains and conditions. During this testing phase, the robot successfully coped with real-world terrain and noise, and maintained strong robustness in the face of perception failures. As shown in Figure, the robot exhibited robust mobility across terrains featuring low friction surfaces, deformable snow, challenging stairs and omnidirectional inclines.

\subsection{Real World Comparison}
Real-world stair experiments demonstrate our method's superior handling of perception uncertainty. As shown in Fig.~\ref{fig:comparison}, the PPO baseline fails due to overconfident foot placement and compounding state estimation errors, leading to unstable transitions and eventual collapse. In contrast, our system achieves reliable ascent through conservative foothold selection with safety margins and full-body stabilization. The world model's temporal filtering effectively mitigates real-world sensor artifacts like radar dropouts and reflection noise, enabling robust navigation where conventional methods fail catastrophically.

\section{CONCLUSIONS AND LIMITATIONS}
In this work, we present Humanoid Perception Controller, a novel framework that synergizes terrain perception with sensor denoising through world model learning and teacher-student distillation. Our experimental results demonstrate three key advantages: (1) The variational information bottleneck enables effective noise suppression while preserving locomotion-critical terrain features, (2) The two-stage distillation paradigm successfully transfers privileged knowledge from oracle to deployable policies, and (3) Integrated perception-action learning outperforms in complex outdoor environments. Extensive real-world validation confirms the stability of the system in realistic uncertain noisy sensing environments.

Our approach nevertheless presents a limitation. The multi-objective learning paradigm introduces complex trade-offs between reconstruction fidelity and policy imitation effectiveness, requiring careful balancing of the variational bottleneck coefficient and imitation loss weight during training to ensure simultaneous convergence of the losses.


%%%%%%%%%%%%%%%%%%%%%%%%%%%%%%%%%%%%%%%%%%%%%%%%%%%%%%%%%%%%%%%%%%%%%%%%%%%%%%%%

%%%%%%%%%%%%%%%%%%%%%%%%%%%%%%%%%%%%%%%%%%%%%%%%%%%%%%%%%%%%%%%%%%%%%%%%%%%%%%%%


% Appendixes

% \section*{ACKNOWLEDGMENT}

% Acknowledge

\bibliography{IEEEfull,references}

\begin{thebibliography}{10}
\providecommand{\url}[1]{#1}
\csname url@rmstyle\endcsname
\providecommand{\newblock}{\relax}
\providecommand{\bibinfo}[2]{#2}
\providecommand\BIBentrySTDinterwordspacing{\spaceskip=0pt\relax}
\providecommand\BIBentryALTinterwordstretchfactor{4}
\providecommand\BIBentryALTinterwordspacing{\spaceskip=\fontdimen2\font plus
\BIBentryALTinterwordstretchfactor\fontdimen3\font minus
  \fontdimen4\font\relax}
\providecommand\BIBforeignlanguage[2]{{%
\expandafter\ifx\csname l@#1\endcsname\relax
\typeout{** WARNING: IEEEtran.bst: No hyphenation pattern has been}%
\typeout{** loaded for the language `#1'. Using the pattern for}%
\typeout{** the default language instead.}%
\else
\language=\csname l@#1\endcsname
\fi
#2}}

\bibitem{RealHumanoid2023}
I.~Radosavovic, T.~Xiao, B.~Zhang, T.~Darrell, J.~Malik, and K.~Sreenath,
  ``Real-world humanoid locomotion with reinforcement learning,''
  \emph{arXiv:2303.03381}, 2023.

\bibitem{siekmann2021blind}
J.~Siekmann, K.~Green, J.~Warila, A.~Fern, and J.~Hurst, ``Blind bipedal stair
  traversal via sim-to-real reinforcement learning,'' \emph{arXiv preprint
  arXiv:2105.08328}, 2021.

\bibitem{gu2024humanoid}
X.~Gu, Y.-J. Wang, and J.~Chen, ``Humanoid-gym: Reinforcement learning for
  humanoid robot with zero-shot sim2real transfer,'' \emph{arXiv preprint
  arXiv:2404.05695}, 2024.

\bibitem{zhang2024whole}
Q.~Zhang, P.~Cui, D.~Yan, J.~Sun, Y.~Duan, G.~Han, W.~Zhao, W.~Zhang, Y.~Guo,
  A.~Zhang, \emph{et~al.}, ``Whole-body humanoid robot locomotion with human
  reference,'' \emph{arXiv preprint arXiv:2402.18294}, 2024.

\bibitem{van2024revisiting}
B.~van Marum, A.~Shrestha, H.~Duan, P.~Dugar, J.~Dao, and A.~Fern, ``Revisiting
  reward design and evaluation for robust humanoid standing and walking,''
  \emph{arXiv preprint arXiv:2404.19173}, 2024.

\bibitem{li2024reinforcement}
Z.~Li, X.~B. Peng, P.~Abbeel, S.~Levine, G.~Berseth, and K.~Sreenath,
  ``Reinforcement learning for versatile, dynamic, and robust bipedal
  locomotion control,'' \emph{The International Journal of Robotics Research},
  p. 02783649241285161, 2024.

\bibitem{HumanoidTerrain2024}
I.~Radosavovic, S.~Kamat, T.~Darrell, and J.~Malik, ``Learning humanoid
  locomotion over challenging terrain,'' \emph{arXiv:2410.03654}, 2024.

\bibitem{radosavovic2024humanoid}
I.~Radosavovic, B.~Zhang, B.~Shi, J.~Rajasegaran, S.~Kamat, T.~Darrell,
  K.~Sreenath, and J.~Malik, ``Humanoid locomotion as next token prediction,''
  \emph{arXiv preprint arXiv:2402.19469}, 2024.

\bibitem{krishna2022linear}
L.~Krishna, G.~A. Castillo, U.~A. Mishra, A.~Hereid, and S.~Kolathaya, ``Linear
  policies are sufficient to realize robust bipedal walking on challenging
  terrains,'' \emph{IEEE Robotics and Automation Letters}, vol.~7, no.~2, pp.
  2047--2054, 2022.

\bibitem{Sun2025LearningHL}
W.~Sun, L.~Chen, Y.~Su, B.~Cao, Y.~Liu, and Z.~Xie, ``Learning humanoid
  locomotion with world model reconstruction,'' \emph{arXiv preprint
  arXiv:2502.16230}, 2025.

\bibitem{xue2025unified}
Y.~Xue, W.~Dong, M.~Liu, W.~Zhang, and J.~Pang, ``A unified and general
  humanoid whole-body controller for fine-grained locomotion,'' \emph{arXiv
  preprint arXiv:2502.03206}, 2025.

\bibitem{long2024learninghumanoidlocomotionperceptive}
J.~Long, J.~Ren, M.~Shi, Z.~Wang, T.~Huang, P.~Luo, and J.~Pang, ``Learning
  humanoid locomotion with perceptive internal model,'' 2024.

\bibitem{wang2025beamdojo}
H.~Wang, Z.~Wang, J.~Ren, Q.~Ben, T.~Huang, W.~Zhang, and J.~Pang, ``Beamdojo:
  Learning agile humanoid locomotion on sparse footholds,'' \emph{arXiv
  preprint arXiv:2502.10363}, 2025.

\bibitem{gu2024advancing}
X.~Gu, Y.-J. Wang, X.~Zhu, C.~Shi, Y.~Guo, Y.~Liu, and J.~Chen, ``Advancing
  humanoid locomotion: Mastering challenging terrains with denoising world
  model learning,'' \emph{arXiv preprint arXiv:2408.14472}, 2024.

\bibitem{miki2022learning}
T.~Miki, J.~Lee, J.~Hwangbo, L.~Wellhausen, V.~Koltun, and M.~Hutter,
  ``Learning robust perceptive locomotion for quadrupedal robots in the wild,''
  \emph{Science robotics}, vol.~7, no.~62, p. eabk2822, 2022.

\bibitem{ha2018world}
D.~Ha and J.~Schmidhuber, ``World models,'' \emph{arXiv preprint
  arXiv:1803.10122}, 2018.

\bibitem{nahrendra2023dreamwaq}
I.~M.~A. Nahrendra, B.~Yu, and H.~Myung, ``Dreamwaq: Learning robust
  quadrupedal locomotion with implicit terrain imagination via deep
  reinforcement learning,'' in \emph{2023 IEEE International Conference on
  Robotics and Automation (ICRA)}.\hskip 1em plus 0.5em minus 0.4em\relax IEEE,
  2023, pp. 5078--5084.

\bibitem{choi2023learning}
S.~Choi, G.~Ji, J.~Park, H.~Kim, J.~Mun, J.~H. Lee, and J.~Hwangbo, ``Learning
  quadrupedal locomotion on deformable terrain,'' \emph{Science Robotics},
  vol.~8, no.~74, p. eade2256, 2023.

\bibitem{kuindersma2016optimization}
S.~Kuindersma, R.~Deits, M.~Fallon, A.~Valenzuela, H.~Dai, F.~Permenter,
  T.~Koolen, P.~Marion, and R.~Tedrake, ``Optimization-based locomotion
  planning, estimation, and control design for the atlas humanoid robot,''
  \emph{Autonomous robots}, vol.~40, pp. 429--455, 2016.

\bibitem{scianca2020mpc}
N.~Scianca, D.~De~Simone, L.~Lanari, and G.~Oriolo, ``Mpc for humanoid gait
  generation: Stability and feasibility,'' \emph{IEEE Transactions on
  Robotics}, vol.~36, no.~4, pp. 1171--1188, 2020.

\bibitem{ishihara2020mpc}
K.~Ishihara and J.~Morimoto, ``Mpc for humanoid control,'' in \emph{Robotics
  Retrospectives-Workshop at RSS 2020}, 2020.

\bibitem{miki2022elevation}
T.~Miki, L.~Wellhausen, R.~Grandia, F.~Jenelten, T.~Homberger, and M.~Hutter,
  ``Elevation mapping for locomotion and navigation using gpu,'' in \emph{2022
  IEEE/RSJ International Conference on Intelligent Robots and Systems
  (IROS)}.\hskip 1em plus 0.5em minus 0.4em\relax IEEE, 2022, pp. 2273--2280.

\bibitem{long2024learning}
J.~Long, J.~Ren, M.~Shi, Z.~Wang, T.~Huang, P.~Luo, and J.~Pang, ``Learning
  humanoid locomotion with perceptive internal model,'' \emph{arXiv preprint
  arXiv:2411.14386}, 2024.

\bibitem{liang2018gpu}
J.~Liang, V.~Makoviychuk, A.~Handa, N.~Chentanez, M.~Macklin, and D.~Fox,
  ``Gpu-accelerated robotic simulation for distributed reinforcement
  learning,'' in \emph{Conference on Robot Learning}.\hskip 1em plus 0.5em
  minus 0.4em\relax PMLR, 2018, pp. 270--282.

\bibitem{todorov2012mujoco}
E.~Todorov, T.~Erez, and Y.~Tassa, ``Mujoco: A physics engine for model-based
  control,'' in \emph{2012 IEEE/RSJ international conference on intelligent
  robots and systems}.\hskip 1em plus 0.5em minus 0.4em\relax IEEE, 2012, pp.
  5026--5033.

\bibitem{Genesis}
\BIBentryALTinterwordspacing
G.~Authors, ``Genesis: A universal and generative physics engine for robotics
  and beyond,'' December 2024. [Online]. Available:
  \url{https://github.com/Genesis-Embodied-AI/Genesis}
\BIBentrySTDinterwordspacing

\bibitem{jenelten2024dtc}
F.~Jenelten, J.~He, F.~Farshidian, and M.~Hutter, ``Dtc: Deep tracking
  control,'' \emph{Science Robotics}, vol.~9, no.~86, p. eadh5401, 2024.

\bibitem{lee2020learning}
J.~Lee, J.~Hwangbo, L.~Wellhausen, V.~Koltun, and M.~Hutter, ``Learning
  quadrupedal locomotion over challenging terrain,'' \emph{Science robotics},
  vol.~5, no.~47, p. eabc5986, 2020.

\bibitem{kumar2021rma}
A.~Kumar, Z.~Fu, D.~Pathak, and J.~Malik, ``Rma: Rapid motor adaptation for
  legged robots,'' \emph{arXiv preprint arXiv:2107.04034}, 2021.

\bibitem{kumar2022adapting}
A.~Kumar, Z.~Li, J.~Zeng, D.~Pathak, K.~Sreenath, and J.~Malik, ``Adapting
  rapid motor adaptation for bipedal robots,'' in \emph{2022 IEEE/RSJ
  International Conference on Intelligent Robots and Systems (IROS)}.\hskip 1em
  plus 0.5em minus 0.4em\relax IEEE, 2022, pp. 1161--1168.

\bibitem{peng2021amp}
X.~B. Peng, Z.~Ma, P.~Abbeel, S.~Levine, and A.~Kanazawa, ``Amp: Adversarial
  motion priors for stylized physics-based character control,'' \emph{ACM
  Trans. Graph.}, vol.~40, no.~4, July 2021.

\bibitem{zhuang2024humanoid}
Z.~Zhuang, S.~Yao, and H.~Zhao, ``Humanoid parkour learning,'' \emph{arXiv
  preprint arXiv:2406.10759}, 2024.

\bibitem{kingma2013auto}
D.~P. Kingma, M.~Welling, \emph{et~al.}, ``Auto-encoding variational bayes,''
  2013.

\bibitem{ross2011reduction}
S.~Ross, G.~Gordon, and D.~Bagnell, ``A reduction of imitation learning and
  structured prediction to no-regret online learning,'' in \emph{Proceedings of
  the fourteenth international conference on artificial intelligence and
  statistics}.\hskip 1em plus 0.5em minus 0.4em\relax JMLR Workshop and
  Conference Proceedings, 2011, pp. 627--635.

\bibitem{mittal2023orbit}
M.~Mittal, C.~Yu, Q.~Yu, J.~Liu, N.~Rudin, D.~Hoeller, J.~L. Yuan, R.~Singh,
  Y.~Guo, H.~Mazhar, A.~Mandlekar, B.~Babich, G.~State, M.~Hutter, and A.~Garg,
  ``Orbit: A unified simulation framework for interactive robot learning
  environments,'' \emph{IEEE Robotics and Automation Letters}, vol.~8, no.~6,
  pp. 3740--3747, 2023.

\bibitem{xu2022fast}
W.~Xu, Y.~Cai, D.~He, J.~Lin, and F.~Zhang, ``Fast-lio2: Fast direct
  lidar-inertial odometry,'' \emph{IEEE Transactions on Robotics}, vol.~38,
  no.~4, pp. 2053--2073, 2022.

\end{thebibliography}

\end{document}